\documentclass[letterpaper, 10 pt, conference]{ieeeconf}  

\usepackage[utf8]{inputenc} 
\usepackage[T1]{fontenc}    
\usepackage{hyperref}       
\usepackage{url}            
\usepackage{booktabs}       
\usepackage{amsfonts}       
\usepackage{nicefrac}       
\usepackage{microtype}      
\usepackage{xcolor}         
\usepackage{multirow}
\usepackage{amsmath}
\usepackage{algorithm}
\usepackage{algpseudocode}
\usepackage{algorithmicx}
\usepackage[capitalise, nameinlink]{cleveref}
\usepackage{graphicx}
\usepackage{cuted}

\usepackage[normalem]{ulem}
\definecolor[named]{JennGreen}{HTML}{21D19F}
\definecolor[named]{TODORed}{HTML}{A30B37}

\newcommand{\algname}{Scanford}
\newcommand{\algabbr}{Scanford}
\newcommand{\fullalg}{Scanford}

\IEEEoverridecommandlockouts                              

\title{\LARGE \bf
Robot-Powered Data Flywheels: Deploying Robots in the Wild for Continual Data Collection and Foundation Model Adaptation
}

\author{Jennifer Grannen$^{1}$, Michelle Pan$^{1}$, Kenneth Llontop$^{1}$, Cherie Ho$^{1}$, \\ Mark Zolotas$^{2}$, Jeannette Bohg$^{1}$, Dorsa Sadigh$^{1}$
\thanks{The authors are with $^{1}$Stanford University, and $^{2}$Toyota Research Institute. \texttt{jgrannen@stanford.edu}}
}

\begin{document}

\maketitle

\vspace{-0.1cm}
\begin{strip}
    \centering
    \vspace{-1.5cm}
    \includegraphics[width=\textwidth]{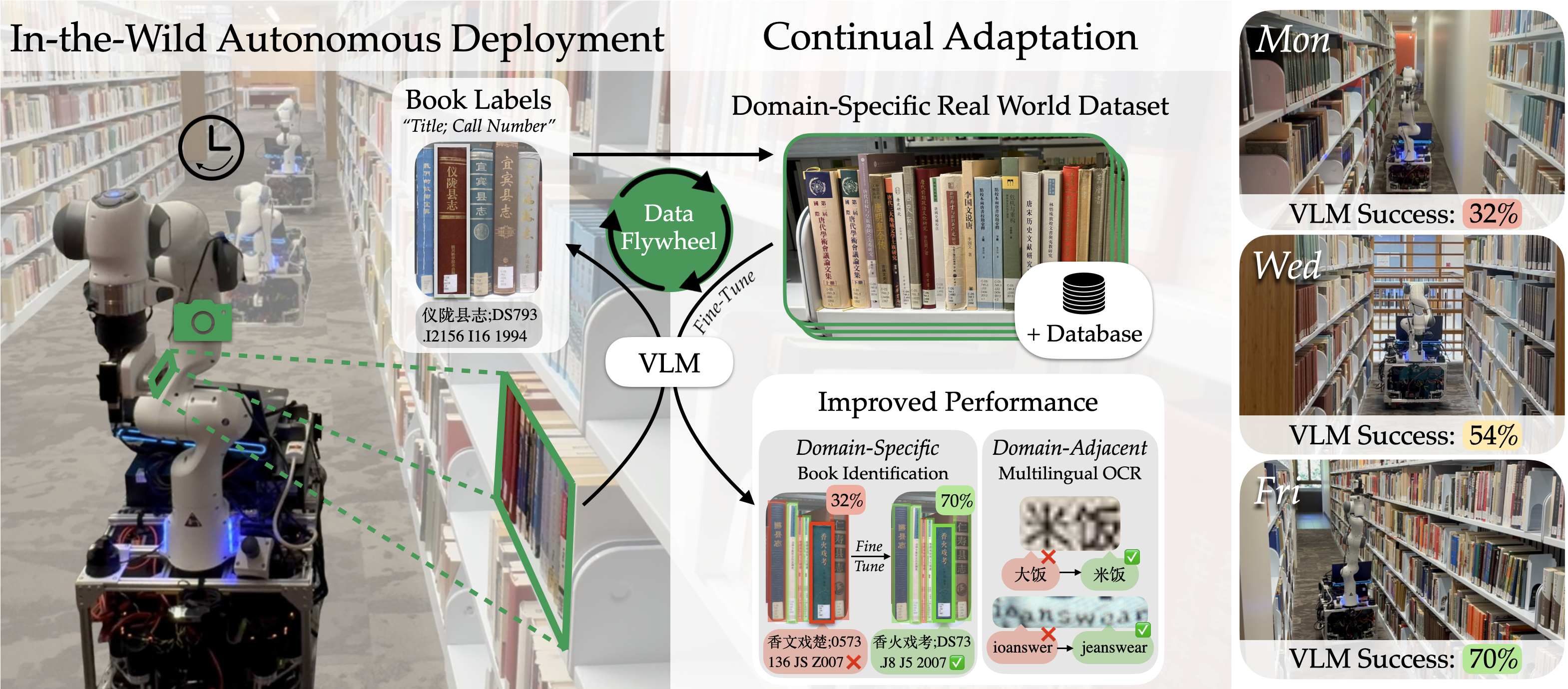}
    \refstepcounter{figure} 
    \vspace{-0.3cm}
    \begin{minipage}{1.0\textwidth}
    \footnotesize
    \setlength{\parskip}{0pt} 
    \setlength{\parindent}{0pt} 
    Fig.~\thefigure. \textbf{\fullalg{}: a Robot-Powered Data Flywheel system for continual foundation model improvement through \textit{in-the-wild} deployment.}
    We deploy \algabbr{} in the East Asia Library for two weeks to scan books and assist inventory management, a challenging setting for off-the-shelf foundation models due to multilingual text, degraded labels, and occlusions [\textbf{Left}]. \algabbr{} uses a mobile manipulator to collect pictures of bookshelves and leverages a VLM to identify the books in each image by title and call number. These labels are then compared with a library catalog database to curate a clean, accurate dataset for VLM fine-tuning [\textbf{Center}].
    Crucially, the autonomously gathered data improves not only the domain-specific performance on book identification, but also domain-adjacent generalizability of foundation models (multilingual OCR).
    \algabbr{} simultaneously (1) saves 18.7 hours of manual scanning, (2) collects real-world book data, and (3) improves the very foundation model it relies on -- enhancing its own performance on the library task while also strengthening the model’s broader multilingual OCR capabilities [\textbf{Right}].
    \end{minipage}
    \label{fig:fig1}
\end{strip}

\thispagestyle{empty}
\pagestyle{empty}

\begin{abstract}


Foundation models have unlocked powerful zero-shot capabilities in vision and language, yet their reliance on internet-sourced pretraining data leaves them brittle in unstructured, real-world environments. The messy, real-world data encountered during deployment -- such as low resolution images, occluded signs, or multilingual text -- remains massively underrepresented in existing corpora.
Robots, as embodied agents, are uniquely positioned to close this gap: they can act in physical environments to collect large-scale, real-world data that enriches foundation model training with precisely the examples current models lack.
We introduce the Robot-Powered Data Flywheel, a framework that transforms robots from consumers of foundation models into data generators. By deploying robots equipped with foundation models in the wild, we enable a virtuous cycle: robots perform useful tasks while simultaneously collecting domain-representative data that improves both domain-specific adaptation and domain-adjacent generalization. 
We instantiate this framework with \fullalg{}, a mobile manipulator robot deployed in the East Asia Library for two weeks. \algabbr{} autonomously scans shelves, identifies books using a vision-language model (VLM), and leverages the library catalog to automatically label images without human annotation. This deployment both aids librarians and produces a curated dataset to finetune the underlying VLM, improving performance on the domain-specific in-the-wild library setting and on domain-adjacent multilingual OCR benchmarks.
Using data collected from 2103 shelves, \algabbr{} improves VLM performance on multilingual book identification from 32.0\% to 71.8\% and boosts domain-adjacent multilingual OCR from 24.8\% to 46.6\% (English) and 30.8\% to 38.0\% (Chinese), while saving an estimated 18.7 hours of human labor. These results highlight how robot-powered data flywheels can both reduce human effort in real deployments and unlock new pathways for continually adapting foundation models to the messy realities of the world. Additional videos and details are available at our website: 
\href{https://scanford-robot.github.io}{https://scanford-robot.github.io}.

\end{abstract}




\section{Introduction}


Recent advances in foundation models have unlocked impressive zero-shot capabilities in vision and language, from optical character recognition (OCR) to image captioning~\cite{comanici2025gemini25pushingfrontier, qwen2025qwen25technicalreport}. However, these systems rely heavily on a vast amount of internet data that are clean, curated, and biased toward certain languages and domains. As a result, they often fail on the ``final mile'' of perception in unstructured environments (e.g., reading nutrition facts on a crumpled wrapper, interpreting graffiti-covered road signs, or identifying book titles on worn bindings of library books)~\cite{chen2021benchmarking,baek2019STRcomparisons}.

The core gap is that the messiness of real-world environments is massively underrepresented in existing pretraining data. Robots, as mobile embodied agents, are uniquely positioned to close this gap: they can autonomously collect large-scale, real-world data directly from the environments where foundation models are ultimately deployed, enriching training corpora with precisely the kinds of real-world examples these models lack.

Our key insight is to transform robots from \textit{consumers} of foundation models into \textit{data generators} that drive a \textbf{Robot-Powered Data Flywheel} (RPDF). By deploying robots equipped with foundation models in-the-wild, we enable them to perform useful tasks while simultaneously collecting domain-representative data that can be used to continually refine those very same foundation models. Crucially, because this data captures domains missing from internet-scale pretraining corpora, it enhances not only \textit{domain-specific} adaptation but also strengthens the foundation model’s broader capabilities in \textit{domain-adjacent} settings (e.g., reading text in low-resolution or occlusion-heavy images). This creates a virtuous cycle where deployment improves models, and improved models enable more successful deployments. 


We instantiate this framework with \textbf{\fullalg{}}, as shown in \cref{fig:fig1}. Inventory management in this library is extremely labor-intensive: at the East Asia Library, it takes five librarians approximately nine months to complete a full cataloging, meaning a complete inventory can only be performed once every 10--15 years. Our robotic system scans shelves with an onboard camera and uses a vision-language model (VLM) to identify the books. Unlike standard benchmarks dominated by English, these shelves contain books primarily in Chinese, Japanese, and Korean -- languages for which current VLMs, and especially OCR capabilities, remain underdeveloped \cite{liu2024mmbench}. 
To close this gap, we automatically label scans using the library's pre-existing catalog and finetune the VLM, improving both domain-specific book recognition and domain-adjacent multilingual OCR. 
This setup highlights the power of a data flywheel: each deployment both leverages existing foundation model capabilities and generates novel, in-the-wild data to improve these very models. Over time, this cycle enables the system to better handle the multilingual, visually cluttered, and domain-specific conditions of this library (\cref{fig:session_shelves}, Bottom), while also improving the robustness and generalization of foundation models to domain-adjacent tasks, in this case multilingual general OCR.

Our contributions are: 
(1) \textbf{Robot-Powered Data Flywheel for Foundation Models}, a framework that leverages robotic deployments to simultaneously perform tasks and collect real-world data, improving both domain-specific and domain-adjacent foundation model capabilities; and
(2) \textbf{\fullalg{}}, an instantiation of the Robot-Powered Data Flywheel that performs in-the-wild book inventory management at a library over an extended deployment (two weeks), showing how embodied settings can yield rich and domain-representative datasets. 
In experiments, we demonstrate the continual adaptation of foundation models -- showing how instantiations of the robot-powered data flywheel framework, such as \algabbr{}, translate deployment into measurable model improvements.
Using our \algabbr{}-collected dataset, we improve model performance on the domain-specific library task from \textbf{32\%} $\rightarrow$ \textbf{71.8\%} and on domain-adjacent multilingual OCR tasks from \textbf{24.8\%} $\rightarrow$ \textbf{46.6\%} (English) and \textbf{30.8\%} $\rightarrow$ \textbf{38.0\%} (Chinese). During this process, we scan more than \textbf{2100} bookshelves, saving a librarian-estimated \textbf{18.7 human hours} compared to manual inventory scanning.

\section{RELATED WORKS}
In this section, we review the state-of-the-art in foundation models and robotics, and describe how \algabbr{} uses these advances to improve foundation models via autonomously-collected real-world data.

\subsection{Foundation Model Generalization and Adaptation}
Foundation models have shown remarkable zero-shot capabilities and generalization across diverse settings and tasks such as OCR, visual question answering, and image captioning~\cite{comanici2025gemini25pushingfrontier, qwen2025qwen25technicalreport, gemmateam2024gemmaopenmodelsbased, touvron2023llamaopenefficientfoundation, Chen2022PaLIAJ, OpenAI_2025_Gpt5}. However, these models can substantially benefit from downstream adaptation~\cite{lee2024vhelmholisticevaluationvision, karamcheti2024prismatic}, especially for capabilities less represented in the pretraining data, such as multilingualism~\cite{SinhaElhafsiEtAl2024,burns2019women,lee2024vhelmholisticevaluationvision}. Such adaptation, often achieved through in-context learning or task-specific finetuning~\cite{singh2019textvqa, Hudson_2019_CVPR}, critically depends on the high quality collection and curation of task-specific datasets~\cite{grattafiori2024llama3herdmodels, hoffman2022computeoptimal}. While internet-scale corpora have powered much of the progress in foundation models, they fall short in a key way: they underrepresent niche or domain-specific settings where robots are often deployed -- i.e., they miss out on the vast amount of real-world data~\cite{liu2024mmbench}.
However, manually collecting such data is both prohibitively time-intensive and logistically difficult in deployment settings.
To address this, we propose leveraging an \textit{autonomous} robot system as a data collection agent, enabling the gathering of domain-specific, real-world data without human effort.
At scale, such autonomous collectors unlock access to vast, previously inaccessible data, capturing the richness and diversity absent from internet corpora with minimal human intervention.
    
\subsection{Robot-Powered Adaptation}
Adaptation in robotics has a rich history, spanning reset-free reinforcement learning \cite{rlscale2023rss, yahya2017collective}, learning from human feedback \cite{liu2022robot, hu2025racrobotlearninglonghorizon, shi2024yell, spencer2020interventions}, and self-supervised learning \cite{hansen2021deployment, chen2024learningonthedrive, sivaprakasam2025salon}. With the advent of foundation models and their demonstrated adaptability, there is a growing opportunity to leverage these models for robotics, and prior work has explored adapting foundation models to specific downstream tasks \cite{grannen2024vocal, grannen2025provox, mattamala25wild, sivaprakasam2025salon, zhai2024finetuning}. While these prior works show clear gains in task-specific performance, they do not explore how the data collected could be used to expand a model’s generalization and reasoning capabilities more broadly beyond the given task.
In contrast, we leverage in-the-wild robotic deployments to continually improve a foundation model by expanding dataset coverage into domains that are underrepresented or absent from internet-accessible data, thus enhancing performance not only on domain-specific tasks but also on domain-adjacent tasks and generalization.

\subsection{In-the-Wild Robot Deployment}
There has been significant progress on robot deployments in-the-wild. Many works focus on generalization to new environments or objects, but with relatively simple, short-horizon skills (such as pick and place or door opening) \cite{etukuru2024robot, stone2023openworld, liu2024okrobot, fangandliu2024moka, rlscale2023rss}.
On the other end are highly-specialized, extensively-engineered systems \cite{jenamani2025feast, Triebel2016spencer} that achieve long-term deployment, but are costly to build, have limited generalization to new tasks, and may even require infrastructure modifications such as RFID tagging for inventory management \cite{PALRobotics}.

Robotic systems powered by foundation models are promising for bridging this gap, offering strong generalization capabilities for more complex, longer-horizon tasks \cite{intelligence2025pi05visionlanguageactionmodelopenworld, hu2023robofm, Kawaharazuka16092024, mattamala25wild}. However, these capabilities typically remain static during deployment. In contrast, our approach continually adapts the model -- improving performance on the specific task at hand while also enhancing generalization by unlocking new domains of real-world data.

Motivated by this gap, we present a robotic system that leverages the general reasoning capabilities of VLMs to operate in a real-world library and kickstart a data flywheel. Our approach demonstrates how robots can adapt a foundation model to improve performance on a domain-specific task (identifying Chinese library books) as well as enhance generalization to domain-adjacent settings such as multilingual OCR. 

\section{Robot-Powered Data Flywheel}


We propose the Robot-Powered Data Flywheel framework as an iterative data collection and adaptation problem. The goal is to continually improve a foundation model (FM) by leveraging a robot deployed in the wild to collect real world data while performing a useful task. Critically, this real world data improves the performance of not only domain-specific tasks but also domain-adjacent tasks.

Let $\text{FM}_0$ denote a pretrained foundation model. We define a discrete set of iterations $t = 1,2,\dots$ corresponding to different deployment cycles, each with a max time horizon of $T$ corresponding to limiting factors such as robot battery life. At iteration $t$, the robot, powered by $\text{FM}_{t-1}$, observes the environment and collects a raw dataset $D^{\text{raw}}_t$.



After collecting the raw dataset $D^{\text{raw}}_t$, 
we apply a curation process to produce a set of final labels. This curation can either be a manual or automated process (we describe an automated instantiation of curation in \cref{sec:library_curation}). Formally, this curation can be viewed as a function $\texttt{Curate}: D^{\text{raw}}_t \mapsto D_t$,
which maps the raw dataset $D^{\text{raw}}_t$ to a curated subset $D_t$. 
This step ensures that only high-quality, reliable labels are retained for inclusion in the training dataset.

We define a dataset $\mathcal{D}_t$ that is accumulated over iterations with the curated data,
\[
\mathcal{D}_t = \bigcup_{k=1}^t D_k, \quad \mathcal{D}_0 = \emptyset,
\]
and used to fine-tune $FM_0$, producing an updated model $\text{FM}_t$ for the next deployment cycle.

In summary, the \algabbr{} implements a closed loop:
\[
D_t^\text{raw} \xrightarrow{\text{Curate}} D_t \xrightarrow{\text{Aggregate}} \mathcal{D}_t \xrightarrow{\text{Fine-tune}} \text{FM}_t \xrightarrow{\text{Robot}} D_{t+1}^\text{raw} 
\]
where robot deployment, data curation, and model adaptation operate iteratively to improve both task-specific and general capabilities of the foundation model.

\begin{algorithm}[t]
\caption{Robot-Powered Data Flywheel}
\begin{algorithmic}[1]

\State Initialize dataset $\mathcal{D}_0 = \emptyset$, pretrained foundation model $\text{FM}_0$, max deployment time $T$
\For{$t = 1,2,\dots$ until task completed}
    \State $D^{\text{raw}}_t \gets \text{RobotDeploy}(\text{FM}_{t-1}, T)$
    \Comment{Robot system collects raw data}
    \State $D_t \gets \texttt{Curate}(D^{\text{raw}}_t)$
    \Comment{Curate and clean data}
    \State $\mathcal{D}_t \gets \mathcal{D}_{t-1} \cup D_t$
    \State $\text{FM}_t \gets \text{FineTune}(\text{FM}_0, \mathcal{D}_t)$
    \Comment{Update FM}
\EndFor


\end{algorithmic}
\label{alg:data_flywheel_library}
\end{algorithm}
\begin{figure*}[t]
    \centering
    \includegraphics[width=\linewidth]{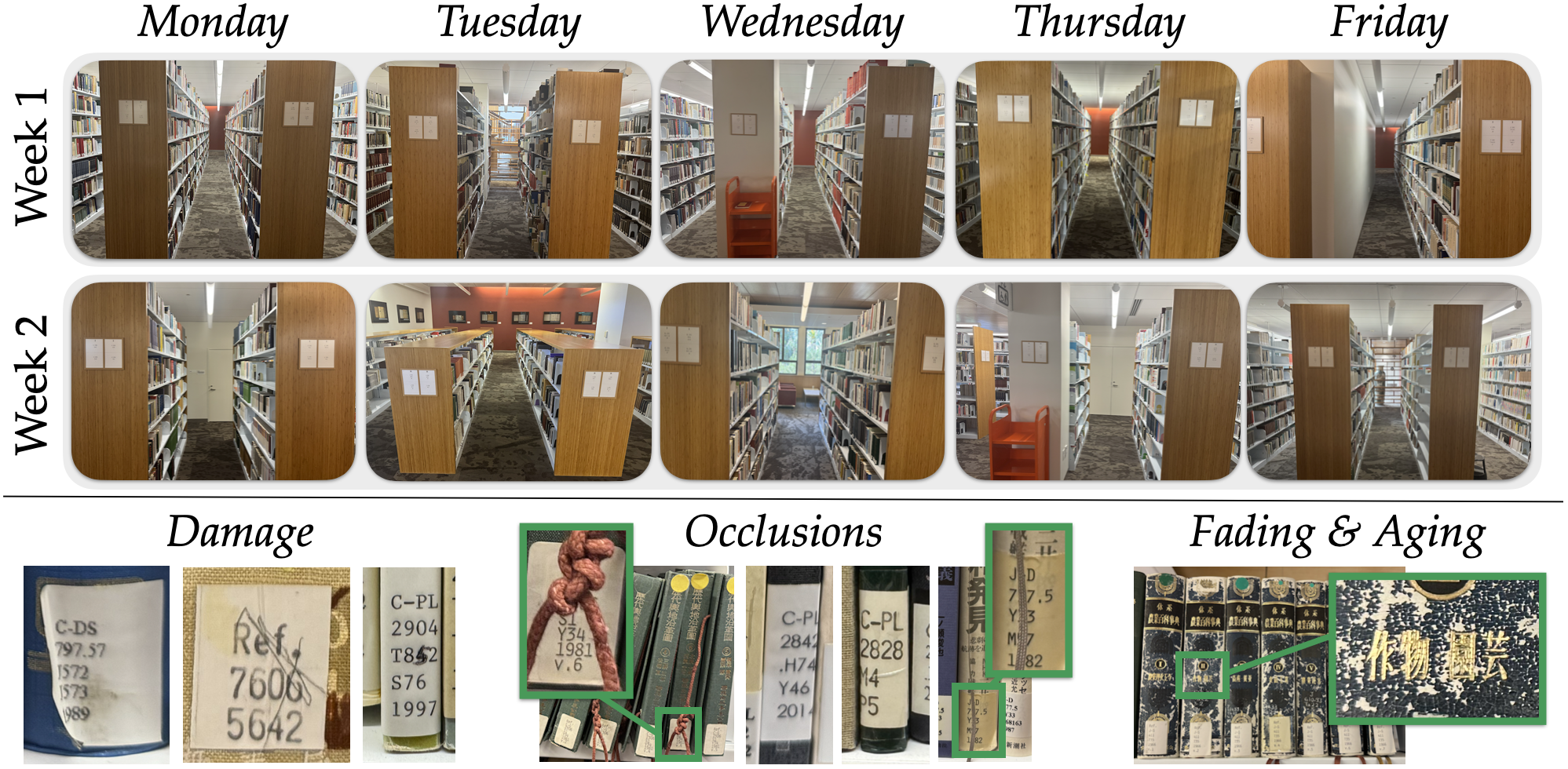}
    \vspace*{-5mm}
    \caption{\textbf{In-the-Wild Challenges at the East Asia Library}: 
    [Top]: We visualize a representative section of library shelves that were scanned for each day of deployment. We note the highly varied setting, with challenges such as varied shelf heights, lengths, backgrounds, and lighting. We especially highlight the exceedingly short shelves on Tuesday of Week 2, which are only three shelves tall rather than the standard height of seven. We hypothesize that this change in height made sensing the shelf positions from the LiDAR data more noisy, leading to a higher number of human interventions needed that day (12). [Bottom]: We present challenges encountered in library shelves (damage, occlusions, and fading/aging of multilingual book labels) which necessitate deploying the data flywheel to improve the VLM performance.}
    \label{fig:session_shelves}
    \vspace*{-3mm}
\end{figure*}

\section{\algname{}}

In this section, we describe the \fullalg{} system, our instantiation of the Robot-Powered Data Flywheel framework. 

We deploy \algabbr{} in the East Asia Library, where shelves contain Chinese, Japanese, and Korean books, each marked with an alphanumeric call number label that uniquely identifies the book (\cref{fig:fig1}). Inventory management -- scanning and reading shelves to determine which books are present or missing -- is a time-intensive task for librarians. Automating this task is particularly challenging: books may be damaged, faded, aged, or partially occluded (see \cref{fig:session_shelves}, Bottom). Moreover, because foundation models are predominantly trained on English-rich internet corpora, multilingual titles further strain their recognition capabilities. 
To address these challenges, we deploy \algabbr{}, a robotic system that leverages a VLM to scan shelves and identify books -- thus simultaneously collecting real-world data. This data is then used to iteratively improve the VLM that powers the system’s perception and reasoning capabilities. 

In the \algabbr{} instantiation, we define a dataset 
\[ D_t = \{(I_t^{(n)}, L_t^{(n)})\}_{n=1}^{N_t}, \quad  I_t^{(n)} \in \mathbb{R}^{H \times W \times 3}, \quad L_t^{(n)} \in \mathcal{L}
\]
where each image $I_t^{(n)}$ represents an RGB observation of height $H$ and width $W$ and has a corresponding label $L_t^{(n)}$. $\mathcal{L}$ denotes the set of all possible strings, in this case the title and call number of all books in the image. $N_t$ is the number of images collected on day $t$ and $D_t^{\text{raw}} = \{(I_t^{(n)}, \hat{L}_t^{(n)})\}_{n=1}^{N_t}$, where $\hat{L}_t^{(n)}$ is the predicted, uncurated label for $I_t^{(n)}$.

\subsection{Shelf Inventory Scanning System}

To scan entire shelf ranges (shown in \cref{fig:session_shelves}, Top), the robot must first be mobile along the floor to traverse the library aisles, while also being able to move vertically to capture both the highest and lowest shelves. To satisfy these requirements, we use a Franka FR3 arm~\cite{FR3} mounted on a TidyBot++ mobile base~\cite{wu2024tidybot}, which is well-suited for library aisles that are at least 36 inches wide, complying with the Americans with Disabilities Act (ADA) standards, given the 21-inch width of the TidyBot base.
The combined platform provides horizontal mobility for aisle traversal and vertical reach for column scanning. The system is equipped with two sensors: a wrist-mounted Intel RealSense D435 RGB-D camera for capturing shelf images $I_t$, and a base-mounted Unitree L2 LiDAR for navigation and localization (see \cref{fig:fig1}, Left).

\smallskip 

\noindent \textbf{Robot Control.} The shelf scanner is programmed with a set of predefined heights corresponding to each shelf level within a column. At each stop along the aisle, the arm moves through these heights to capture images $i_t$ of all shelves, then the base advances $0.3\,\text{m}$—approximately half the image coverage—before repeating the process.

In practice, odometry alone is unreliable due to wheel drift. Traditional SLAM methods fail in the narrow, visually homogeneous environment of book aisles, where repetitive geometry prevents robust localization. To address this, we leverage raw LiDAR point cloud data for heuristic drift correction. Specifically, the shelves on either side of the aisle appear as two dense vertical clusters in the point cloud. By fitting planes to these clusters, the robot can estimate the aisle boundaries, then correct its position to remain centered and align its heading parallel to the shelves. This drift correction is performed before each vertical scan, ensuring consistent coverage along the aisle.

\smallskip 

\noindent \textbf{Data Labeling.} After each scanning session, we associate the collected images with the corresponding library section, which maps to a known call number range according to the Library of Congress cataloging system \cite{congressclassification}. This provides a natural ordering of books on each shelf and allows us to narrow the candidate set of books for each image to those listed in the database for the corresponding section. While the catalog is useful for proposing a candidate set of books, it alone cannot open-loop label the images as it is not 100\% accurate: books may be checked out, in use elsewhere, or stored offsite. 

Instead, we use a VLM $\text{FM}_{t-1}$ to predict more accurate labels for each image. Let $D_t^{\text{raw}} = \{(I_t^{(n)}, \hat{L}_t^{(n)})\}_{n=1}^{N_t}$ denote the set of images and raw label predictions collected at robot deployment iteration $t$, where each $I_t^{(n)} \in \mathbb{R}^{810 \times 1080 \times 3}$ is an RGB image of a shelf segment, and $\hat{L}_t^{(n)} \in \mathcal{L}$ is the corresponding predicted label: the titles and call numbers of all the books in the image from left to right. 
To avoid spurious model outputs that do not correspond to candidate books, we use a retrieval-augmented generation (RAG)-based approach \cite{gao2024retrievalaugmentedgenerationlargelanguage} that passes the set of candidate books from the database entries as context in the VLM prompt. 
This step increases the likelihood that its predictions align with the correct subset of books for the shelf section.  

Together, the robot’s control system and the foundation model comprise the full robotic shelf scanning system, enabling end-to-end collection and labeling of all books along the library aisles while performing the inventory task.

\subsection{Data Curation}
\label{sec:library_curation}

After collecting a set of raw shelf data, we must clean and curate the data for an accurate training dataset. Given a set of raw data $D_t^{\text{raw}}$ collected by the robot at iteration $t$, our goal is to construct a curated dataset of $D_t = \{(I_t^{(n)}, L_t^{(n)})\}_{n=1}^{N_t}$ pairs, where $L_t^{(n)}$ are the verified labels for each image.  

Thus, we instantiate a library-specific autonomous curation function $\texttt{Curate} : (D_t^{\text{raw}}) \mapsto D_t,$ that leverages the library database's ordered candidate books. 
For each predicted set of book labels, we want to confirm that (1) each book is a valid book within the set of candidate books and (2) the ordered list of books matches the ordering in the catalog.
Thus, validation is performed using (1) string similarity matching between predicted books and database entries~\cite{Ratcliff1998Gestalt}, and (2) local ordering checks to ensure neighboring books in the prediction are consistent with the database.  

Image–label pairs that exceed a threshold similarity score are accepted into the dataset, while others are discarded. This semi-automatic process accounts for the fact that raw predictions from the VLM are imperfect, enabling scalable, high-quality labeling without requiring manual annotation. 
We provide detailed statistics on the number of accepted versus rejected pairs and dataset sizes in \cref{sec:experiments}.

\subsection{Model Adaptation}

Having constructed the curated dataset, we leverage it to improve the performance of the foundation model, closing the loop of the data flywheel. At each iteration $t$, we fine-tune a pretrained VLM $\text{FM}_0$ on the accumulated dataset $\mathcal{D}_t$, after which newly collected data from subsequent deployments are incorporated and the process is repeated. This iterative procedure enables the model to continually adapt to the domain-specific task of multilingual book identification while also improving performance on domain-adjacent tasks.

\smallskip

\noindent \textbf{Training Details.}
We instantiate $\text{FM}$ with Qwen2.5-VL (7B), chosen for its open-source availability and demonstrated multilingual capabilities (English and Chinese) \cite{qwen2025qwen25technicalreport}.
We fine-tune the pretrained model for 5 epochs on each dataset using a single Nvidia H200 GPU, with training times ranging from 1 to 7 hours depending on dataset size (487–5,019 images).
Fine-tuning is performed using the AdamW optimizer with a learning rate of $2 \cdot 10^{-7}$. We use a per-device batch size of 4 with gradient accumulation of 4 steps, resulting in an effective batch size of 16. Mixed-precision training with \texttt{bfloat16} is employed, along with weight decay of 0.01 and a cosine learning rate scheduler with a 3\% warmup ratio.

\section{EXPERIMENTS}
\label{sec:experiments}

We evaluate the \algname{} on two fronts: (1) domain-specific and domain-adjacent adaptation of foundation models, and (2) task deployment success.

\begin{figure*}[t]
    \centering
    \includegraphics[width=\linewidth]{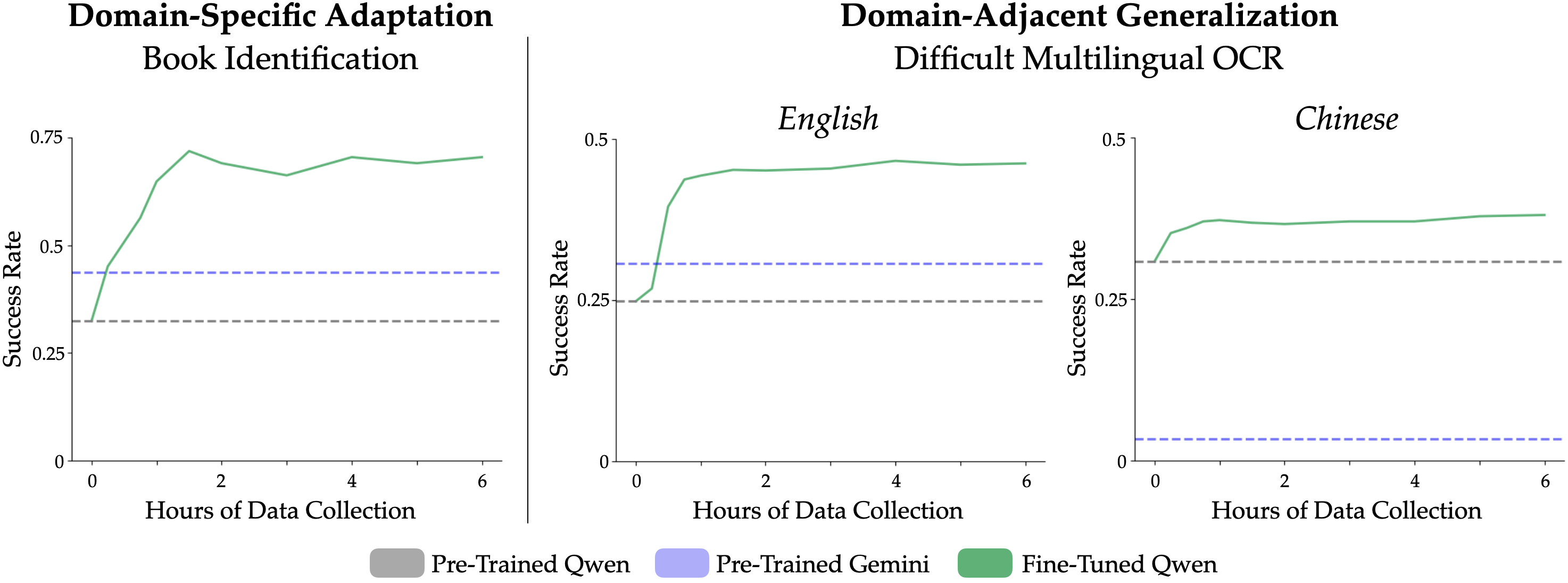}
    \vspace*{-5mm}
    \caption{\textbf{Adaptation Results with Flywheel Data: Fine-tuning a VLM improves both domain-specific performance and domain-adjacent generalization.} Fine-tuning Qwen2.5 for book identification increased performance from 32.4\% to 71.8\% [\textbf{Left}]. 
    Critically, fine-tuning Qwen with the same data also achieves impressive gains for domain-adjacent generalization (multilingual OCR) -- from 24.8\% to 46.6\% for English text and from 30.8\% to 38.0\% for Chinese text [\textbf{Right}]. We hypothesize Gemini's poor performance at Chinese OCR is because it likely has less Chinese in its pretaining mixture. Qwen on the other hand emphasizes Chinese performance during its pretraining \cite{qwen2025qwen25technicalreport}. }
    \label{fig:adaptation_results}
    \vspace*{-3mm}
\end{figure*}

\subsection{Model Adaptation}
\label{sec:adaptation_results}

A central insight of the Robot-Powered Data Flywheel for Foundation Models is that data collected during deployment can be repurposed to improve foundation models -- not only for the deployed task itself but also for closely related settings. 
In \algabbr{}, the robot collects images of library shelves along with labels for book titles and call numbers. 
After 6 hours of deployment, this raw dataset has 8,232 labeled images, and is curated by filtering with the library's database to remove images with incorrectly predicted labels (as described in \cref{sec:library_curation}). 
We use this final curated dataset of 5,019 labeled images to improve performance in two scenarios: (1) the domain-specific task of identifying books in shelf images, and (2) the domain-adjacent task of multilingual OCR on challenging datasets with low resolution or occlusion.

\subsubsection{Library Book Identification (Domain-Specific)}
\cref{fig:adaptation_results}, Left visualizes the results of domain-specific adaptation. We finetune Qwen2.5~\cite{qwen2025qwen25technicalreport} on the curated data collected during our shelf scanning robot deployment and evaluate model performance on a held-out, hand-labeled set of 71 images, corresponding to 10 shelves. While pretrained models such as Qwen2.5 and Gemini achieve moderate accuracy (32.4\% and 43.7\% respectively), their performance is limited by the challenges of in-the-wild deployment, including book damage, fading, and frequent occlusions (\cref{fig:session_shelves}, Bottom), motivating the need for fine-tuning on real-world data.
After finetuning on our collected data, Qwen2.5 reaches 71.8\%, a 39.4\% absolute improvement over its pretrained baseline. Interestingly, most of this gain is achieved within the first 1.5 hours of deployment ($\approx$ 1,352 images), after which gains plateau. This finding highlights that even short deployments can yield data that substantially improves VLM performance, directly enhancing downstream task success.

\subsubsection{Multilingual OCR (Domain-Adjacent)}

While domain-specific adaptation demonstrates clear utility, the domain-adjacent results highlight what truly sets the Robot-Powered Data Flywheel apart: the ability of robot-collected data to improve a model’s broader, non-task-specific capabilities. \cref{fig:adaptation_results}, Right, shows multilingual OCR performance before and after finetuning with \algabbr{} deployment data.

We evaluate two OCR benchmarks: (1) an English dataset \cite{baek2019STRcomparisons} (644 images) and (2) a Chinese dataset \cite{chen2021benchmarking} (500 images). To stress-test generalization, we focus on each benchmark's most difficult subset (by their categorization \cite{baek2019STRcomparisons, chen2021benchmarking}) -- failure cases in English (heavy occlusions, low resolution, calligraphic fonts, vertical text) and ``hard'' cases in Chinese (occlusion, background clutter, blur, curved or vertical text). These settings more closely mirror the noisy, real-world conditions faced during deployment.

Our results show that robot-collected library data improves Qwen2.5’s ability to generalize to these challenging scenarios. For English OCR, finetuned Qwen2.5 reaches 46.6\% accuracy, compared to only 24.8\% and 30.7\% for pretrained Qwen2.5 and Gemini. Chinese OCR is more difficult due to the large set of visually similar characters \cite{chen2021benchmarking}, yet finetuning still yields significant gains: pretrained Qwen2.5 and Gemini achieve 30.8\% and 3.4\%, respectively, while finetuned Qwen2.5 improves to 38.0\%. As with the domain-specific case, these gains plateau after roughly 1.5 hours of deployment ($\approx$ 1,352 images). We hypothesize that Gemini's poor performance at ``challenging'' Chinese OCR is likely because it has less Chinese in its pretraining mixture relative to Qwen2.5~\cite{qwen2025qwen25technicalreport}, coupled with the complexity and visual similarity of Chinese characters~\cite{chen2021benchmarking}.

These findings support our claim that underrepresentation of certain domains in internet-scale pretraining corpora (e.g., low-quality OCR) leads to systematic weaknesses in pretrained models. The Robot-Powered Data Flywheel framework addresses this gap by collecting precisely such in-the-wild data, broadening coverage and driving measurable improvements in robustness and generalization.

\subsection{Library Shelf Scanning and Data Collection}

\begin{figure*}[t]
    \centering
    \includegraphics[width=\linewidth]{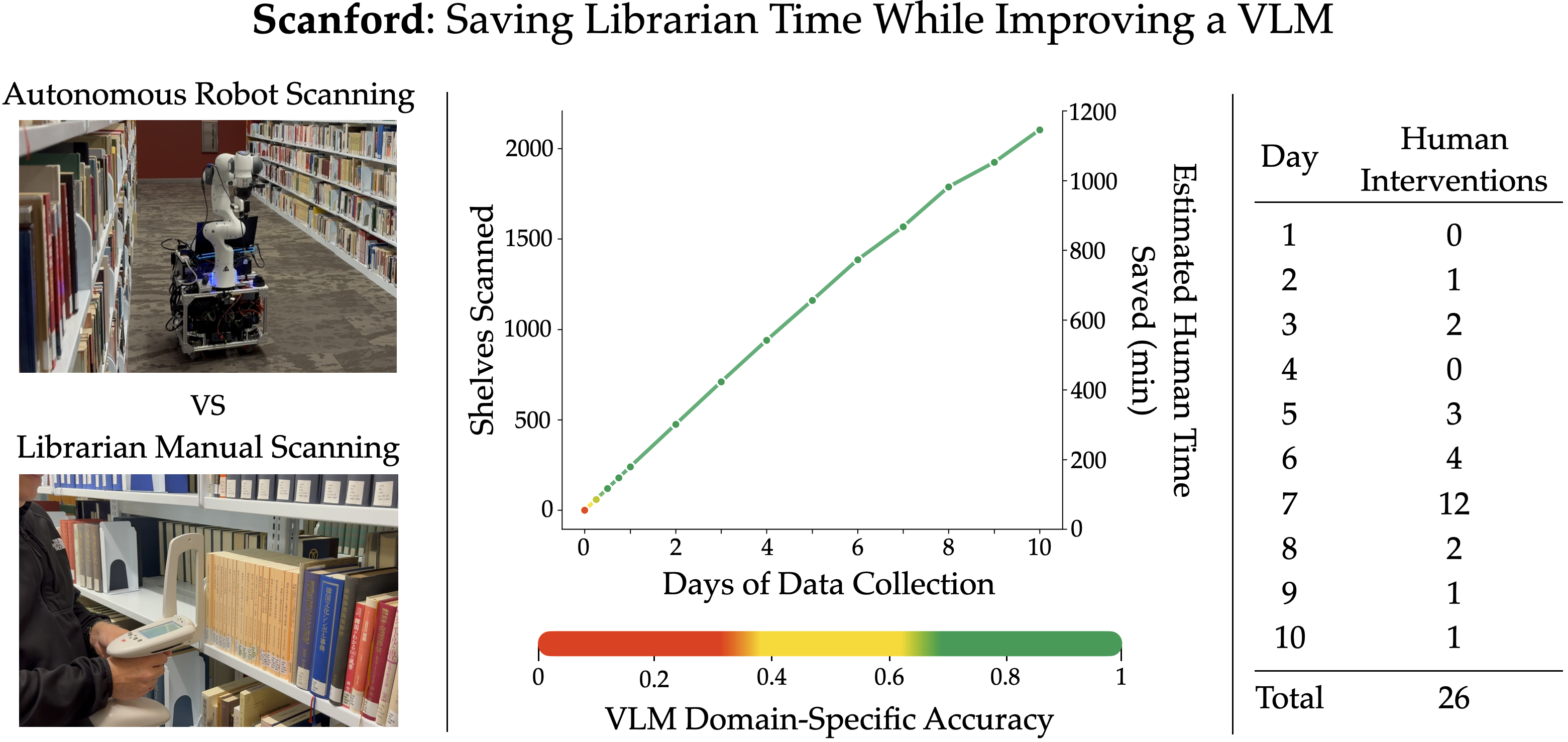}
    \vspace*{-5mm}
    \caption{\textbf{\algname{} saves librarian time while improving VLM performance}. We report the results of \algabbr{}'s autonomous, in-the-wild deployment in the East Asia Library. \algabbr{} scanned and labeled 2,103 shelves -- saving a librarian's estimate of 18.7 hours of human time [\textbf{Left}] -- while simultaneously improving VLM performance [\textbf{Center}]. Over 10 days of data collection, only 26 human interventions were needed, each averaging under 5 minutes [\textbf{Right}].}
    \label{fig:deployment_results}
    \vspace*{-3mm}
\end{figure*}

\algabbr{} is not just a data collection system -- it also performs a valuable real-world task: library shelf scanning for inventory management. We evaluate its effectiveness in \cref{fig:deployment_results}. Over two weeks of deployment (10 days, 4 hours per day), \algabbr{} scanned a total of 2,103 shelves, a workload that librarians estimate would otherwise require 18.7 hours of manual effort.
At the same time, the system improved its own capabilities by adapting its VLM component, as illustrated by the performance gains shown in \cref{fig:deployment_results}, Center (see \cref{sec:adaptation_results} for details). Across the full deployment, \algabbr{} required only 26 human interventions -- an average of 2.6 per day -- with each intervention lasting under 5 minutes to correct the robot's drift and recenter it within the aisle.  
These results highlight \algabbr{}’s dual benefits: reducing human labor while continually improving the foundation model that powers it.







\section{DISCUSSION}

We introduce the \textbf{Robot-Powered Data Flywheel} (RPDF), a framework that leverages robotic deployments to both accomplish useful tasks and collect real-world data that enhances foundation model capabilities. By fine-tuning on messy, in-the-wild data, we address domain gaps that are poorly represented in internet pretraining data, boosting foundation model performance in both domain-specific and domain-adjacent settings.
We instantiate this framework with \textbf{\fullalg{}}, a mobile manipulator deployed at the East Asia Library to perform book inventory management. This setting poses challenges rarely captured in VLM pretraining data, including multilingual book titles, visually cluttered shelves, and damaged or faded text -- all of which result in poor out-of-the-box performance. By fine-tuning a VLM on data collected during deployment, we improve accuracy on the library-specific book identification task from \textbf{32.0\%} $\rightarrow$ \textbf{71.8\%}. Importantly, these domain-specific gains also transfer to domain-adjacent tasks: the VLM’s general OCR performance improves from \textbf{24.8\%} $\rightarrow$ \textbf{46.6\%} in English and from \textbf{30.8\%} $\rightarrow$ \textbf{38.0\%} in Chinese, highlighting the broader benefits of real-world data collection. Beyond data generation, \algabbr{}’s \textbf{two-week deployment} (40 hours) also provides immediate utility: by scanning more than \textbf{2100} bookshelves, it saves an estimated \textbf{18.7 hours} of librarian effort compared to manual inventory scanning.


We are excited for the community to build upon the RPDF framework to create flywheels across diverse applications, ultimately enabling large-scale data collection with little human effort. Promising applications for similar VLM improvements include grocery stores (occluded packaging or calligraphic fonts) or hospitals (handwritten prescriptions and expiry dates). From our experience designing \algabbr{}, we highlight several insights that may help guide future instantiations:

\smallskip

\noindent \textbf{Task Selection.} 
When designing new instantiations of the RPDF, we draw an analogy to pedagogy: deployment settings should lie within the \textit{Zone of Proximal Development}~\cite{Vygotsky1978} of current foundation model and robotic capabilities. This ensures that the raw data collected remains both useful and tractable for model improvement. If the task is too easy (e.g., navigating by reading large-font signage), there is little room for the model to improve, limiting the benefits of the flywheel. On the other hand, if the task is too difficult (e.g., manipulating and reading foreign newspapers), the data will require extensive manual labeling, curation, or robot intervention, undermining the autonomy and scalability of the flywheel.

\smallskip

\noindent \textbf{Deploy in unstructured settings.}
The greatest benefits emerge in messy, real-world domains where internet corpora are underrepresented. By targeting these “frontier” environments, robotic deployments can uncover valuable, novel data distributions that meaningfully extend model capabilities.

\smallskip

\noindent \textbf{Stakeholder alignment.}
Finally, effective flywheels require selecting tasks that are not only tractable and informative for models, but also directly helpful to human stakeholders. This ensures that robotic deployments create immediate utility while  generating data to improve future deployments.

\smallskip

\noindent \textbf{Limitations and Future Work.} While the RPDF framework is broadly applicable across deployments and foundation model classes, our current instantiation, \algabbr{}, only considers a VLM and requires moderate task-specific engineering to operate effectively. Moreover, despite significant performance gains, we do not achieve 100\% success on either the domain-specific or domain-adjacent tasks, suggesting that fine-tuning alone may be insufficient to guarantee full robustness.

Looking forward, we aim to extend this framework to other types of foundation models, such as large language models (LLMs)~\cite{OpenAI_2025_Gpt5} and vision-language-action models (VLAs)~\cite{intelligence2025pi05visionlanguageactionmodelopenworld,kim24openvla}, to explore how in-the-wild deployments can address underrepresented aspects of their pretraining data.
We expect this to be especially impactful for VLAs, where robotics data is far more limited than in LLMs or VLMs. 
Leveraging the embodied reasoning capabilities of these foundation models for robot control will also reduce the engineering effort needed for each deployment. 

We are also interested in exploring alternative ways of incorporating deployment-collected data -- such as integrating it into the pretraining mixture or improving data curation strategies to ensure accurate fine-tuning data -- with the goal of approaching 100\% task success. Beyond adaptation, data gathered through Robot-Powered Data Flywheels could be compiled into a richer evaluation benchmark, better reflecting the messy and diverse conditions of real-world environments.

\section{Acknowledgments}
\footnotesize
\noindent \textit{AI Tool Disclosure}: The authors are responsible for all content in this article. AI tools (Claude, ChatGPT) were used in a limited capacity (minor grammar enhancement and code autocompletion). 
\normalsize









\bibliographystyle{IEEEtran}
\bibliography{references}

\end{document}